\documentclass[letterpaper, 10 pt, conference]{ieeeconf}  

\IEEEoverridecommandlockouts                              
\usepackage[utf8]{inputenc}
\usepackage[T1]{fontenc}
\usepackage[dvipsnames]{xcolor}
\usepackage{graphics} 
\usepackage{amsmath} 
\usepackage{amssymb}  
\usepackage[final]{pdfpages}
\usepackage{booktabs}
\usepackage{changepage}
\usepackage{ragged2e}
\usepackage{comment}
\usepackage{siunitx}
\usepackage{hyperref}

\title{\LARGE \bf
Mitigating Compensatory Movements in Prosthesis Users\\ via Adaptive Collaborative Robotics
}

\author{Marta Lagomarsino$^1$, Robin Arbaud$^{1,2}$, Francesco Tassi$^1$, and Arash Ajoudani$^1$
\thanks{This work was supported by PR FESR 2021–2027: Incentivi alle imprese per attività collaborativa di ricerca industriale e sviluppo sperimentale. Bando DGR 2026/2021. Progetto RE-FINGER: Prat. n. 2022/38.}
\thanks{{$^1$Human-Robot Interfaces and Interaction, Istituto Italiano di Tecnologia, Genoa, Italy.} {$^2$ Dept. of Informatics, Bioengineering, Robotics and System Engineering, University of Genoa, Genoa, Italy.}}
\thanks{Corresponding author's email: {\tt\footnotesize marta.lagomarsino@iit.it}}
}

\begin{document}

\maketitle
\thispagestyle{empty}
\pagestyle{empty}

\begin{abstract}
Prosthesis users can regain partial limb functionality, however, full natural limb mobility is rarely restored, often resulting in compensatory movements that lead to discomfort, inefficiency, and long-term physical strain.
To address this issue, we propose a novel human-robot collaboration framework to mitigate compensatory mechanisms in upper-limb prosthesis users by exploiting their residual motion capabilities while respecting task requirements. Our approach introduces a personalised mobility model that quantifies joint-specific functional limitations and the cost of compensatory movements. This model is integrated into a constrained optimisation framework that computes optimal user postures for task performance, balancing functionality and comfort. The solution guides a collaborative robot to reconfigure the task environment, promoting effective interaction.
We validated the framework using a new body-powered prosthetic device for single-finger amputation, which enhances grasping capabilities through synergistic closure with the hand but imposes wrist constraints. Initial experiments with healthy subjects wearing the prosthesis as a supernumerary finger demonstrated that a robotic assistant embedding the user-specific mobility model outperformed human partners in handover tasks, improving both the efficiency of the prosthesis user's grasp and reducing compensatory movements in functioning joints. These results highlight the potential of collaborative robots as effective workplace and caregiving assistants, promoting inclusion and better integration of prosthetic devices into daily tasks. 
\end{abstract}


\section{Introduction}

The loss of a body limb or digit profoundly impacts an individual's physical capabilities and mental well-being. 
Prosthetic devices aim to mitigate these issues by restoring functionality and enabling users to regain independence in daily living and work-related activities. Prostheses can be classified into three categories: passive, body-powered, and externally powered \cite{imbinto2016}. 
Passive devices primarily focus on restoring appearance rather than mechanical functionality 
\cite{goiato2009}, however,  they can still help to restore opposing grasp capabilities \cite{Fraser1998}.
Body-powered devices, in contrast, utilise the motion of another muscle or limb to control the prosthesis, offering a mechanical and functional simple solution \cite{carey2015differences, imbinto2016}. 
Externally powered devices rely on actuators, often electric or pneumatic, to restore finer control and dexterity, albeit at the cost of increased complexity, reliance on a power source, and higher expense \cite{milazzo2024design, marinelli2023active}. 
By comparison, body-powered devices strike a balance between functionality and usability due to their simplicity and straightforward use.

Despite their benefits, prostheses remain far from replicating the full functionality of natural limbs \cite{bumbavsirevic2020current}. They often impose constraints on the user's body mobility, leading to compensatory mechanisms that may involve uncomfortable non-ergonomic postures or inefficient movement patterns \cite{carey2008compensatory, metzger2012characterization}. This is especially true for body-powered systems, which by design require some muscles to control the prosthesis in addition to their normal purposes \cite{engdahl2022comparison, valevicius2020compensatory}. Over time, these compensations can result in strain, discomfort, and even long-term physical issues. Unfortunately, such challenges are frequently overlooked in the design and assessment of prosthetic devices. 
To address this gap, studying how human kinematics can effectively adapt to residual impairments and prosthetic limitations provides an opportunity to enhance user comfort, efficiency, and overall experience.

Beyond individual adaptations, the environments where prosthesis users operate (such as workplaces, care facilities, and homes) can be modified to support more ergonomic and functional interactions. 
In this context, collaborative robots (CoBots) constitute a promising solution
\cite{cakmak2023physically}. CoBots have demonstrated their ability to improve physical ergonomics by online adapting interaction poses \cite{Bestick2018Learning} 
or optimising shared kino-dynamics in co-carrying and co-manipulation tasks \cite{Kim2019Adaptable, ferraguti2020unified}. They have also been shown to promote cognitive ergonomics by adapting the proximity and reactivity of the CoBot to balance safety, user stress, and productivity \cite{lagomarsino2024pro}. 
While some initial attempts aimed at facilitating tasks for users with impairments, these approaches often focused on completely substituting the impaired limb functionality with a robotic system \cite{torielli2024laser, poirier2019voice} or limiting the movement of the impaired arm in Cartesian space \cite{Ardon2021affordance}. 
However, such strategies overlook the nuanced needs of prosthetic users who experience joint-level and task-specific constraints. Additionally, these methods fail to exploit the residual functionality of restricted-mobility users, potentially reinforcing a sense of inability and frustration.

This work takes a novel 
step toward addressing these challenges by introducing a human-robot collaboration framework that supports upper-limb prosthesis users in overcoming joint-level limitations and exploiting their residual motion but avoiding harmful compensatory mechanisms. 
We propose a model of joint-specific functional limitations of prosthesis users and quantify the cost of compensatory mechanisms in functioning joints. This model informs an optimisation problem that computes the optimal user posture for performing a collaborative task in a functional and comfortable manner, considering task constraints. The solution 
guides the robot to reconfigure the task environment thus promoting the adoption of the recommended posture.
Finally, we present a novel body-powered prosthetic device for single-finger amputees, designed to enhance grasping capabilities through synergistic hand closure but posing constraints in wrist motion. The device serves as a testbed to validate the proposed method. 

The remainder of the paper is organised as follows: Section \ref{sec:methods} describes the modelling of upper-body impairments and the formulation of the optimisation problem for robot-assisted facilitation to minimise the use of joints with restricted mobility and mitigate the cost associated with compensatory mechanisms. Section \ref{sec:experiments_results} presents the experimental campaign to validate the proposed approach in handover tasks. Section \ref{sec:discussions_conclusions} discusses the results and outlines potential directions for future work.

\section{Methods}
\label{sec:methods}

\begin{figure}[t!]
    \centering
    \includegraphics[width=\linewidth]{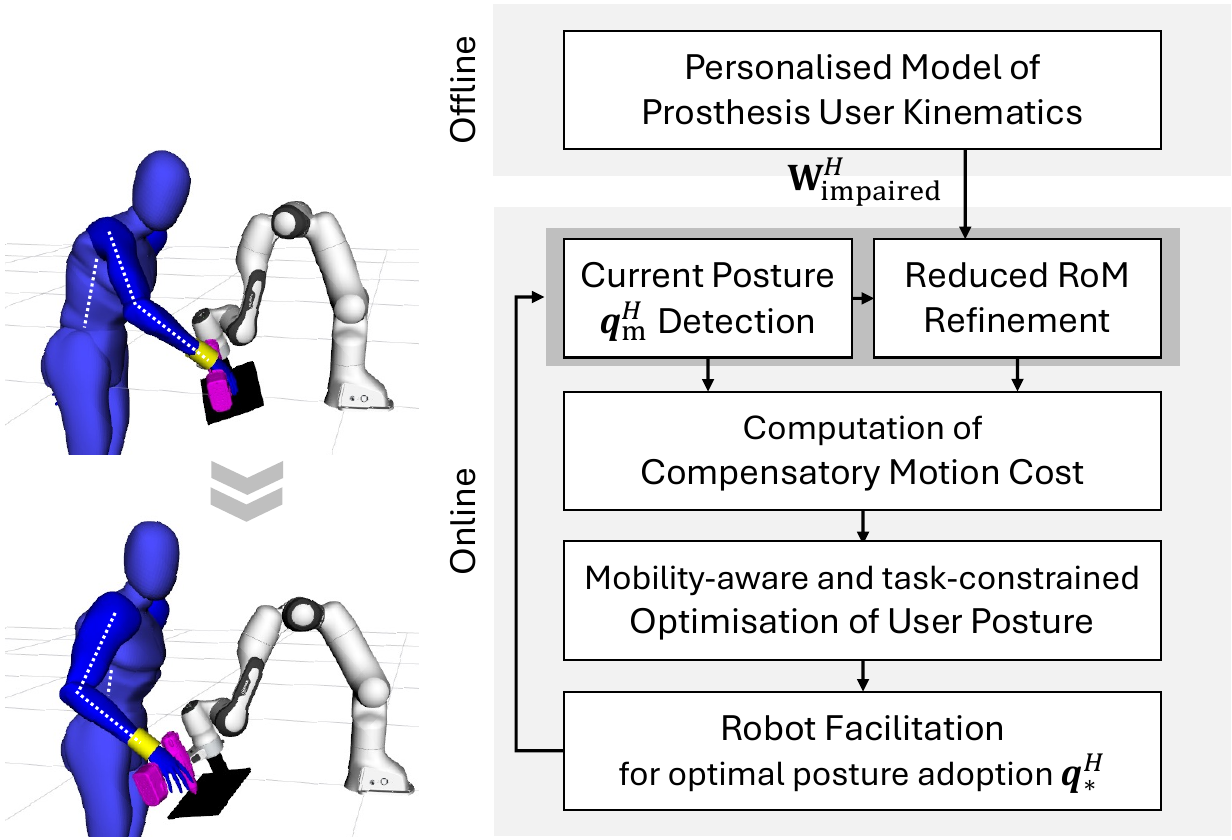}
    \caption{Overview of the procedure to assess the mobility diversity of prosthesis users and mitigate compensatory movements while adhering to task constraints.}
    \label{fig:schema}
\end{figure}

The proposed procedure to mitigate compensatory movements in upper-limb prosthesis users is outlined in Fig. \ref{fig:schema}. In the offline phase, we first model the user mobility diversity and scale a digital model to automatically match their specific body measurements. During the interaction phase, we continuously monitor the prosthesis user's posture and eventually refine the Range of Motion (RoM) based on modelled residual mobility and prosthetic limitations. 
Consequently, we calculate the current cost of compensatory motions, defined as the deviation of the functioning joints from a natural, comfortable posture. A constrained optimisation technique is then solved to determine a more comfortable and functional posture that still respects the user's mobility constraints and task requirements. Finally, a robot-assisted motion is planned to facilitate the adoption of such posture. 

\subsection{Prosthesis User Kinematics Model}
This study models the kinematics of a prosthesis user by assigning a local coordinate frame to each joint, where the x-, y-, and z-axes define the rotation axes for abduction/adduction, flexion/extension, and internal/external rotation, respectively, as outlined in \cite{leonori2023bridge}. To account for arm kinematic redundancy, the arm is represented as a seven Degrees-of-Freedom (DoF) system, denoted by $\boldsymbol{q}^H \in \mathbb{R}^7$ and consists of two rigid links: the upper arm ($\boldsymbol{l}_\text{humerus}$) and the forearm ($\boldsymbol{l}_\text{radius}$) \cite{bertomeu2018human}. 

A coordinate frame is also assigned to the L5 lumbar spine vertebra to represent spine flexion through the first joint angle, $q^H_{1}$, and the link, $\boldsymbol{l}_\text{spine}$.  This inclusion reflects the well-documented role of increased trunk motion as a compensatory mechanism in the literature \cite{valevicius2020compensatory}. The resulting model presents $M=8$ DoFs, where size-related parameters can be adjusted to fit different arm and trunk sizes.

The shoulder is modelled as a spherical joint, providing abduction/adduction ($q^H_{2}$), flexion/extension ($q^H_{3}$), and internal/external rotation ($q^H_{4}$). The elbow consists of both a hinge and a pivot joint that allows for forearm flexion/extension ($q^H_{5}$) and pronation/supination ($q^H_{6}$). Finally, the wrist joint is represented as a condyloid joint, enabling flexion/extension ($q^H_{7}$) and ulnar/radial deviation ($q^H_{8}$) of the hand. Table \ref{tab:dh_params} provides the Denavit-Hartenberg (DH) parameters used for this kinematic model.

\begin{table}[!h]
\vspace{-0.3cm}
    \centering
    \caption{\small Denavit-Hartenberg parameters of \\ the adopted human kinematic model.} 
    \label{tab:dh_params}
    \begin{tabular}{lcccc}
         \toprule
            axis, $i$ & $\boldsymbol{\theta}_{i}$ & $a_i$ & $\alpha_i$ & $d_i$ \\
            \cmidrule[0.4pt](r{0.125em}){1-1}
            \cmidrule[0.4pt](lr{0.125em}){2-5}
            1 & $q_1^H$         & $l_\text{spine}^z$     & --$\frac{\pi}{2}$  & $-l_\text{spine}^y$ \\
            2 & $q_2^H$         & 0                 &  $\frac{\pi}{2}$  & $-l_\text{spine}^x$ \\
            3 & $q_3^H$+$\frac{\pi}{2}$   & 0                 &  $\frac{\pi}{2}$  & 0 \\
            4 & $q_4^H$--$\frac{\pi}{2}$   & -- $l_\text{humerus}^x$  &  $\frac{\pi}{2}$  & -- $l_\text{humerus}^z$ \\
            5 & $q_5^H$+$\pi$     & 0                 &  $\frac{\pi}{2}$  & -- $l_\text{humerus}^y$ \\
            6 & $q_6^H$+$\frac{\pi}{2}$   & -- $l_\text{radius}^y$   &  $\frac{\pi}{2}$  & -- $l_\text{radius}^z$ \\
            7 & $q_7^H$+$\frac{\pi}{2}$   & 0                 &  $\frac{\pi}{2}$  & -- $l_\text{radius}^x$ \\
            hand & $q_8^H$      & 0                 &  0        & 0 \\
            \bottomrule
    \end{tabular}
\end{table}

Prostheses assist users by restoring some functionality, though they may not fully replicate natural limb mobility, and can impose limitations on surrounding joints. For instance, wearing a partial hand prosthesis may pose limitations in wrist movement, which can lead to compensatory movements in the elbow and/or shoulder. To capture these constraints in the model, 
we define a diagonal matrix, $\textbf{W}_\text{impaired}^H \in \mathbb{R}^{M \times M}$. This matrix modifies the kinematic model 
by prioritising movement in healthy or prosthesis-supported joints and reducing or entirely limiting mobility in impaired or blocked joints.
This allows for flexibility in modelling partial or complete impairments across joints. For joints with complete functional loss, we set $\textbf{W}_\text{impaired}^H(i,i) = 1$, whereas partially impaired joints are defined proportionally to reflect a preference for minimal use unless essential.
In clinical practice, standardised impairment indices are used to assess joint functionality in patients with conditions such as arthritis, injuries, or neurological damage \cite{rondinelli2023ama}. These indices can be applied to define specific joint limitations within $\textbf{W}_\text{impaired}^H$, helping to personalise the kinematic model to the particular needs and abilities of each prosthesis user.

The cost $\Psi(k) \in \mathbb{R}$, quantifying the impact of compensatory mechanisms at the time instant $k$, is calculated as: 
\begin{equation}
    \Psi[k] = \left\| \big(\textbf{I}-\textbf{W}^H_{\text{impaired}}\big) \, \big(\boldsymbol{q}^H[k] - \boldsymbol{q}^H_{\text{n}}\big) \right\|^2,
    \label{eq:ergo_cost}
\end{equation}
where $\boldsymbol{q}^H_{\text{n}}$ denotes the natural posture, representing a neutral and comfortable arm/trunk configuration for the user. 
This cost measures the deviation of functioning joints from this natural posture. It aligns with established metrics in the literature, such as "instantaneous joints usage" or "ergonomic cost" \cite{Gholami2022quantitative, Bestick2018Learning}, but is refined here by including weights that account for the desired mobility of each joint, as defined by the matrix $\big(\textbf{I}-\textbf{W}^H_{\text{impaired}}\big)$. This modification provides a more precise and personalised assessment of alignment with a comfortable and functional posture, tailored to the user's specific impairments and requirements.




\subsection{Robot Reactive Facilitation for Prosthesis Users}
The objective of the robot reactive facilitation is to guide prosthesis users in performing the interactive task in a more effective way, avoiding unnatural postures adopted to compensate for constraints imposed or not fully addressed by the prosthesis. 
To achieve this, 
the robot generates movements designed to facilitate human adoption of this recommended configuration. The optimal posture minimises the usage of partially impaired joints, completely restricts motion in blocked joints, and promotes alignment of healthy null-space joints with a comfortable and functional posture throughout the interaction.

The robot reactive behaviour is formulated as an online optimisation problem that considers the user's impaired mobility, kinematic constraints, task requirements, and safety distance thresholds. The optimisation is defined by the following cost function and constraints:
\begin{align}
    \min_{\boldsymbol{q}^H} \, 
    & \left\| \textbf{W}^H_{\text{impaired}} \, (\boldsymbol{q}^H - \boldsymbol{q}^H_{\text{m}}) \right\|^2 \!\!\!+ \alpha \left\| (\textbf{I}-\textbf{W}^H_{\text{impaired}}) \, (\boldsymbol{q}^H - \boldsymbol{q}^H_{\text{n}}) \right\|^2
    \nonumber\\
    & \text{s.t. } \,\,\, \boldsymbol{q}^H \in \big[\boldsymbol{q}_{\text{min}, \text{impaired}}^{H}, \boldsymbol{q}_{\text{max},\text{impaired}}^{H} \big], \nonumber\\
    & \quad \quad \boldsymbol{x}_\text{obj}{(\boldsymbol{q}^{H})} \in \boldsymbol{\chi}^R_\text{task}, \nonumber\\
    & \quad \quad p_\text{obj}{(\boldsymbol{q}^{H})} = p_\text{task}, \nonumber\\
    & \quad \quad d_{\text{obj}}^H \geq d_{\text{safe th}}, \nonumber\\
    & \quad \quad d_{\text{elbow}}^H 
    \geq d_{\text{elbow th}},
    \label{eq:optimisation}
\end{align}
where $\boldsymbol{q}^H_{\text{m}}$ is the current measured human posture and $\boldsymbol{q}^H_{\text{n}}$ denotes the natural and comfortable posture for a healthy individual. In the cost function, the first term aims to minimise the deviation of fully or partially impaired joints from their current configuration. This is implemented through the matrix $\textbf{W}^H_{\text{impaired}}$, which weights the squared distance based on the severity of each joint functional limitation, effectively prioritising the minimisation of displacement for more severely impaired joints. 
The second term instead minimises the cost  $\Psi$ of the compensatory mechanism, namely the squared distance of the functioning joints from the nominal most comfortable arm/trunk configuration, weighted by their desired mobility. 
The parameter $\alpha \in \mathbb{R}$ is a scaling factor that gives less importance to the second term. This reflects the prioritisation of maintaining the current position of impaired joints while subtly promoting the comfortable positioning of the healthy, unconstrained joints. 

The optimisation constraints ensure that the joint angles $\boldsymbol{q}^H$ remain in the feasible impaired range, which is derived by reducing the standard RoM 
to account for user-specific impairments. 
For prosthesis users, this reduction is modelled by modifying the healthy joint boundaries $\boldsymbol{q}_\text{min}^{H}$ and $\boldsymbol{q}_\text{max}^{H}$ using the matrix $\textbf{W}_\text{impaired}^H$, as follows: 
\begin{align}
	\boldsymbol{q}_{\text{min}, \text{impaired}}^{H} & = \boldsymbol{q}_\text{min}^{H} + \textbf{W}_\text{impaired}^H \big((\boldsymbol{q}_\text{m}^{H} - \zeta) - \boldsymbol{q}_\text{min}^H\big), \nonumber \\
    \boldsymbol{q}_{\text{max},\text{impaired}}^{H} & = \boldsymbol{q}_\text{max}^{H} - \textbf{W}_\text{impaired}^H \big(\boldsymbol{q}_\text{max}^H - (\boldsymbol{q}_\text{m}^{H} + \zeta)\big), 
\end{align}
where the parameter $\zeta \in \mathbb{R}$ allows small adjustments in the joint angles, even in severely impaired joints. 
Task-specific constraints ensure that the position $\boldsymbol{x}_\text{obj}$ of the object or tool, which will be in the human's hand during the interaction with the environment ($\boldsymbol{x}_\text{obj}$ = $\boldsymbol{x}_\text{obj}{(\boldsymbol{q}^{H})}$), is located within the task space of the robot $\boldsymbol{\chi}^R_\text{task}$ and satisfies task objectives. In particular, we consider the possibility of imposing an equality constraint on a specific Cartesian coordinate $p_\text{obj} = \{x_\text{obj}; y_\text{obj}; z_\text{obj}\} \in \mathbb{R}$ to match the task requirement $p_\text{task}$. Finally, the horizontal distance $d_{\text{obj}}^H \in \mathbb{R}$ between the object and the user's body must exceed a safety threshold $d_{\text{safe th}}$ and the horizontal distance $d_{\text{elbow}}^H$ between the user's elbow and pelvis is required to exceed $d_{\text{elbow th}}$ to ensure a proper elbow pose.

\subsection{Finger Prosthesis System}

\begin{figure}
    \centering
    \includegraphics[width=0.6\linewidth]{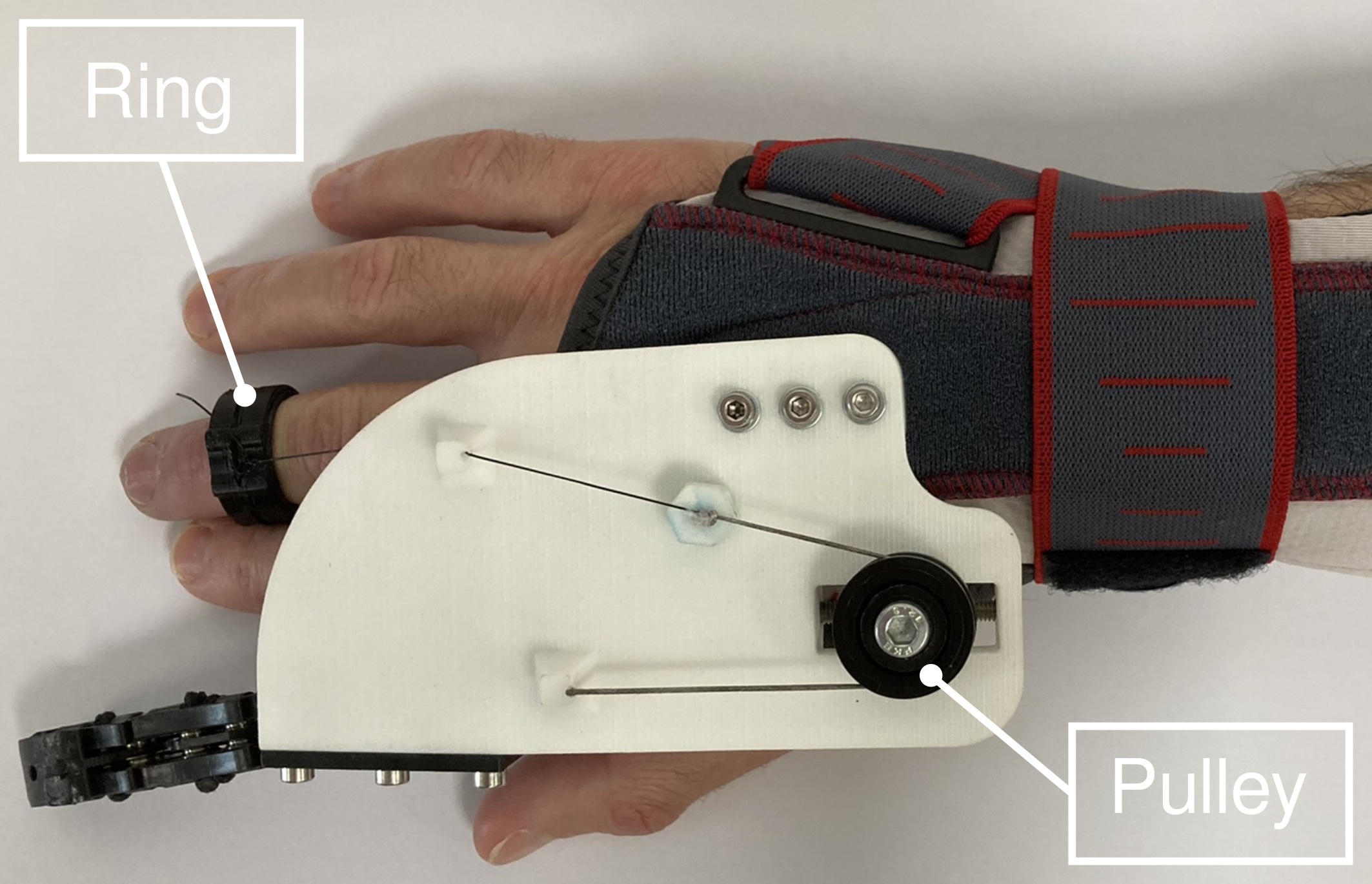}
    \vspace{-0.3cm}
    \caption{Picture of the wearable prototype for testing the proposed body-powered prosthesis as a supernumerary finger.}
    \label{fig:finger_design}
    \vspace{-0.4cm}
\end{figure}

In this work, we introduce a novel body-powered finger prosthetic device specifically designed to enhance grasping for individuals with single-finger amputations, featuring a mechanical coupling mechanism that enables synergistic closure with the hand. 
The prosthesis features an under-actuated $3$-joint finger, whose motion is controlled by a single tendon. Elastic elements bring the finger back into an open position when the tendon is not actively pulled. As illustrated in Fig. \ref{fig:finger_design}, we built a wearable prototype around a wrist brace with a 3D-printed part attached to it. The latter holds a pulley, through which the tendon is rerouted. The other end of the tendon is attached to a ring, which the user has to wear on the finger controlling the prosthetic. Bending that finger will then pull on the tendon and bend the prosthetic. Additionally, a lead screw mechanism can be used to adjust the pretension of the tendon by changing the position of the pulley. 
For testing purposes, we mounted the designed prosthesis as a supernumerary finger placed near the index. This configuration allows non-impaired individuals to simulate the experience of the prosthesis users.

It is important to note that while the prosthesis enhances grasping functionality, it also imposes constraints on the wrist flexion/extension and ulnar/radial deviation. These limitations can be encoded in the $\textbf{W}_\text{impaired}^H$ by setting the last two diagonal elements to $1$, informing the optimisation process of the immobilised wrist.

\section{Experiments}
\label{sec:experiments_results}

\subsection{Experimental Setup and Protocol}
Experiments were conducted with two healthy human subjects to assess the effectiveness of the proposed method in facilitating functional and comfortable interaction with a prosthetic device\footnotemark[1]. The participants included Subject $1$, a $29$-year-old male with a height of $1.83$ \si{m}, and Subject $2$, a $28$-year-old female with a height of $1.58$ \si{m}. Diverse physical profiles were deliberately chosen to validate the method ability to adapt to diverse anthropometric characteristics.

The experimental task required subjects to grasp a hotmelt glue pistol positioned at varying lateral distances from their pelvis, i.e. $y_\text{task} = \{0.05\si{m}$; $-0.20\si{m}$; $-0.45\si{m}\}$. To simulate a quite realistic prosthesis-user scenario, participants performed the task while wearing the proposed prosthetic device configured as a supernumerary finger (see Fig. \ref{fig:finger_design}). 
Before the task, participants underwent a familiarisation phase with the prosthesis.
The setup included a table in front of the prosthesis user, marked with three tape lines indicating the potential lateral directions from which the object could be delivered.
The healthy RoM was derived from the literature \cite{whitmore2012nasa} and the nominal prosthesis user posture, $\boldsymbol{q}^H_\text{n}$, was defined as the most ergonomic posture based on the Rapid Upper Limb Assessment (RULA) Tool \cite{mcatamney2004rula}, with the elbow flexed at $90^\circ$ and all other joint angles set to $0^\circ$. The parameters were set as follows: $\alpha = 0.10$, $\zeta = 5^\circ$, $d_{\text{safe th}}=0.20 \si{m}$ from the human pelvis, and $d_{\text{elbow th}}=0.25 \si{m}$. 

The experiment involved two sessions: human passing (\textit{HP}) and robot passing (\textit{RP}). In the \textit{HP} session, a human participant passed the object to the prosthesis user once for each lateral distance $y_\text{task}$, simulating traditional handover scenarios involving a human co-worker in industrial settings or a caregiver in assistive contexts. Five healthy participants (four males and a female, $28.2 \pm 2.4$ years old) were recruited to serve as object passers in this condition. In the \textit{RP} session, the task was executed in collaboration with a Franka Emika Panda manipulator, controlled at $1$ k\si{Hz}, equipped with the Robotiq 2-Finger Gripper 2F-85. The robot optimised the object transfer pose online by solving the constrained optimisation problem defined in Eq. \eqref{eq:optimisation}, implemented through the Augmented Lagrangian method of the ALGLIB library\footnotemark[2] 
in a C++ environment. 
To ensure smooth execution, a B-spline trajectory was generated to achieve the optimised handover pose, which was tracked by the robot Cartesian impedance controller, as in \cite{lagomarsino2024pro}. 
The robot repeated the object passing at different $y_\text{task}$ five times to ensure a consistent comparison with the \textit{HP} condition.
\footnotetext[1]{
The experiments were carried out at the HRII Laboratory of the Istituto Italiano di Tecnologia in accordance with the Helsinki Declaration, and the protocol was approved by the ethics committee Azienda Sanitaria Locale Genovese N.3 (Protocol IIT\_HRII\_ERGOLEAN 156/2020).}
\footnotetext[2]{\href{https://www.alglib.net/}{https://www.alglib.net/}}

We used a wearable MVN Biomech suit (Xsens Tech.BV) equipped with inertial measurement unit sensors to measure body segment lengths and automatically personalise the kinematic model and digital mannequin meshes for the subject. The motion capture system continuously tracked the prosthesis user's movements throughout the experiment.  
The post-hoc statistical analysis compared the effectiveness of the robot-assisted (\textit{RP}) condition to the baseline human passing (\textit{HP}) condition, with a focus on the impact of embedding a user impairment model in the robot behaviour. Initially, the gathered data were tested for normality using the Anderson-Darling test. If normality was confirmed, repeated measures ANOVA was conducted. In cases where normality was not confirmed, the non-parametric Friedman test was employed to evaluate significant differences. Three performance metrics on the prosthesis user's motion were computed: 
\begin{itemize}
    \item \textit{Task duration} ($T_f$): the time taken by the prosthesis user to complete the approach and grasping phases, measured from the first movement of the grasping subject, until he/she had full control of the object; 
    \item \textit{Functioning joints usage} ($\bar{\Psi}$): calculated during the duration $T_f$ (with time samples $k = 1, \dots, K_f$) as:
    \vspace{-0.1cm}
    \begin{equation}
    \vspace{-0.1cm}
        \bar{\Psi} = \dfrac{1}{K_f} \sum_{k=1}^{K_f} \Psi[k], 
    \end{equation} 
    where $\Psi[k]$ is the cost associated with compensatory mechanisms performed by the functioning joints, as defined in Eq. \eqref{eq:ergo_cost};
    \item \textit{Smoothness of joint motion} ($J$): defined as:
    \vspace{-0.1cm}
    \begin{equation}
    \vspace{-0.1cm}
        J = \Delta t \sum_{k=1}^{K_f} \|\dddot{\boldsymbol{q}}^H_\text{m}[k]\|^2,
    \end{equation}
    where $\Delta t$ is the data sampling rate of the Xsens system and $\dddot{\boldsymbol{q}}^H_\text{m} \in \mathbb{R}^M$ is the joint jerk vector derived numerically from joint velocity measurements. 
\end{itemize}

\subsection{Experimental Results}

\begin{figure}[t]
    \centering
    \includegraphics[width=\linewidth]{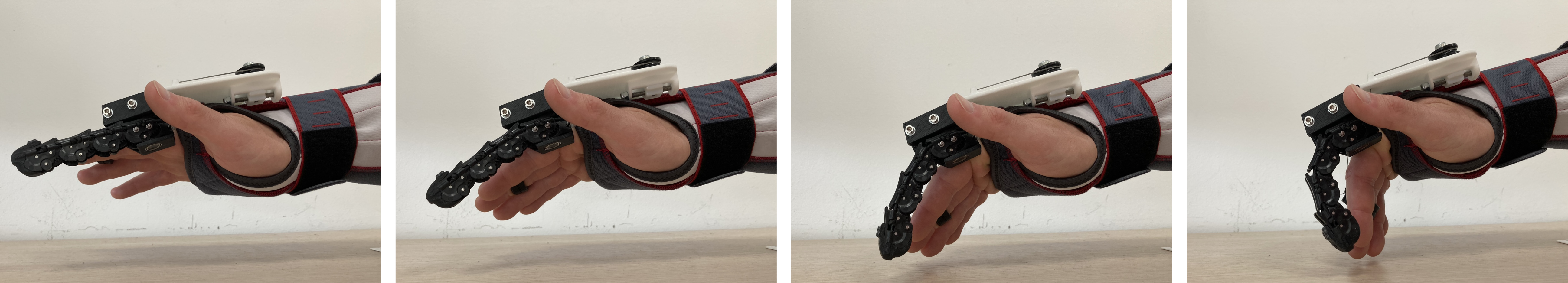}
    \vspace{-0.7cm}
    \caption{Illustration of the coordinated flexion movement of the user's finger and the prosthetic one.}
    \label{fig:finger_movement}
    \vspace{-0.4cm}
\end{figure}

Figure \ref{fig:finger_movement} shows the sequential frames of the middle finger and the supernumerary prosthetic finger in motion to illustrate the progressive flexion movements enabled by the coupling mechanism.

\begin{figure}[b!]
\vspace{-0.5cm}
    \centering
    \includegraphics[width=\linewidth]{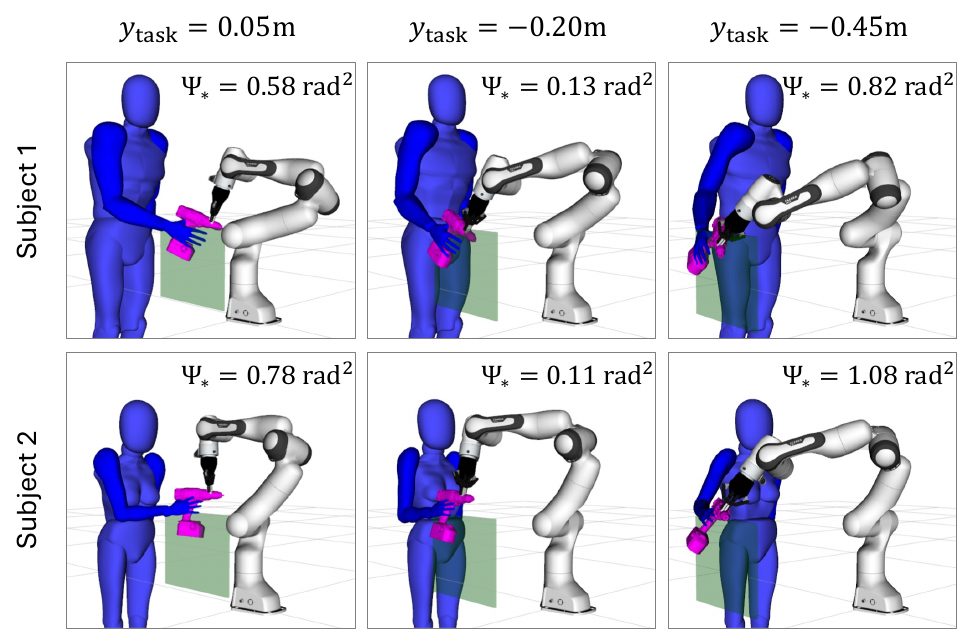}
    \vspace{-0.7cm}
    \caption{Optimised robot-to-prosthesis user handover configurations at different lateral distances from the human pelvis on the table,  tailored to diverse human body measurements.}
    \label{fig:rh_handovers}
\end{figure}

The results of mobility-aware optimisation tailored to specific prosthesis users are reported in Fig. \ref{fig:rh_handovers}, revealing its ability to effectively configure human-robot interaction under varying physical profiles and task constraints. The digital model was automatically scaled to match the anthropometric characteristics of both users based on motion-tracking data. The robot complied with task requirements, namely maintaining different lateral distances from the user's pelvis and adhering to task-specific object positioning areas (indicated by green planes in the figure). More importantly, the robot adapted the position and orientation of the handover object to eliminate wrist movement entirely and mitigate compensatory movements. Each subfigure also reports the cost associated with compensatory mechanisms, $\Psi_*$, for the ideal case where the human assumed the optimal posture derived from the optimisation.

\begin{figure}[!t]
    \centering
    \begin{adjustwidth}{-0.3cm}{0cm}
        
    \includegraphics[width=\linewidth]{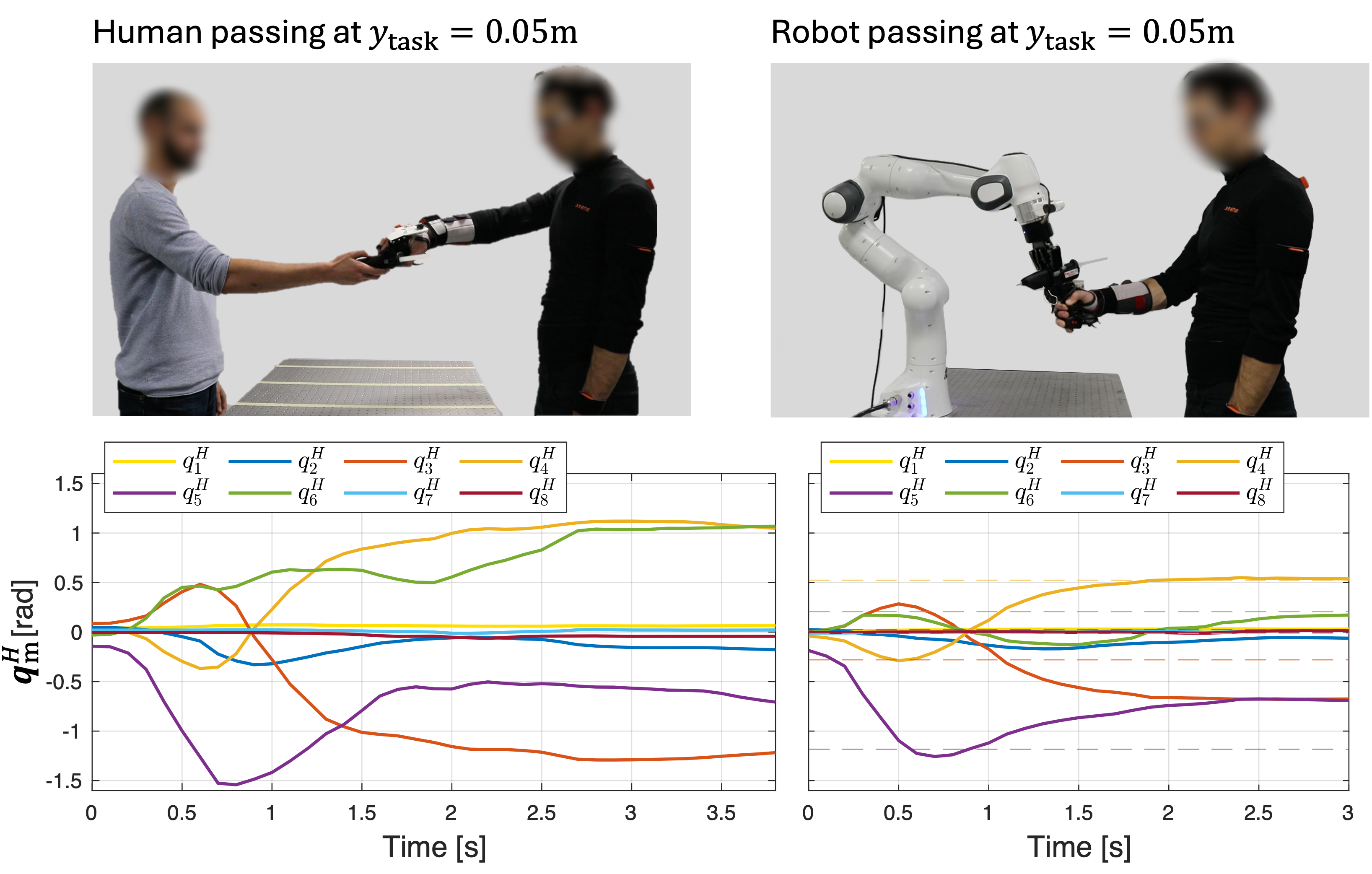}
    \end{adjustwidth}
    \vspace{-0.3cm}
    \caption{Comparison between human-to-prosthesis user handover performed by a participant (\textit{HP}) and optimised robotic handover adapted to the prosthesis user's mobility limitations (\textit{RP}).}
    \label{fig:hh_vs_rh_handover}
\end{figure}

Figure \ref{fig:hh_vs_rh_handover} compares the handover scenarios between human-to-prosthesis user interaction conducted by Participant 3 (\textit{HP}) and the optimised robotic handover (\textit{RP}). Despite the human passer's aim to facilitate the prosthesis user's grasp, the resulting object transfer pose necessitated pronounced compensatory movements from the user due to wrist blockage. Specifically, as shown in the bottom-left plot of Fig. \ref{fig:hh_vs_rh_handover}, the prosthesis user exhibited 
greater elbow flexion (purple line), substantial shoulder flexion (orange line), and an elevated elbow posture, which induced larger shoulder external rotation (yellow line) and forearm pronation (green line). Further, the prosthesis user required approximately a second to stabilise the prosthetic grasp in this configuration. Additionally, a hesitation at the beginning of the movement suggested a slightly increased demand on the neural mechanisms for motion planning, as also reported by the prosthesis user. 
In contrast, the optimised robotic handover, tailored to accommodate the prosthesis user's mobility limitations, enabled a more accessible and efficient grasp. The robot facilitation induces the user to adopt a posture closely aligned with the optimised configuration $\boldsymbol{q}^H_*$ obtained by the mobility-aware optimisation problem (depicted with dashed lines in the bottom-right plot of Fig. \ref{fig:hh_vs_rh_handover}). While the optimised posture recommended greater reliance on elbow flexion and reduced shoulder flexion, the user adopted a configuration with more balanced flexion between the elbow and shoulder.
Despite these minor deviations due to the user's slightly different resolution of the redundancy, the optimised robotic facilitation notably reduced the compensatory movements cost $\Psi$ both at the interaction pose and throughout the approach phase compared to the \textit{HP} condition. 

\begin{figure}[!t]
    \centering
    \includegraphics[width=0.8\linewidth]{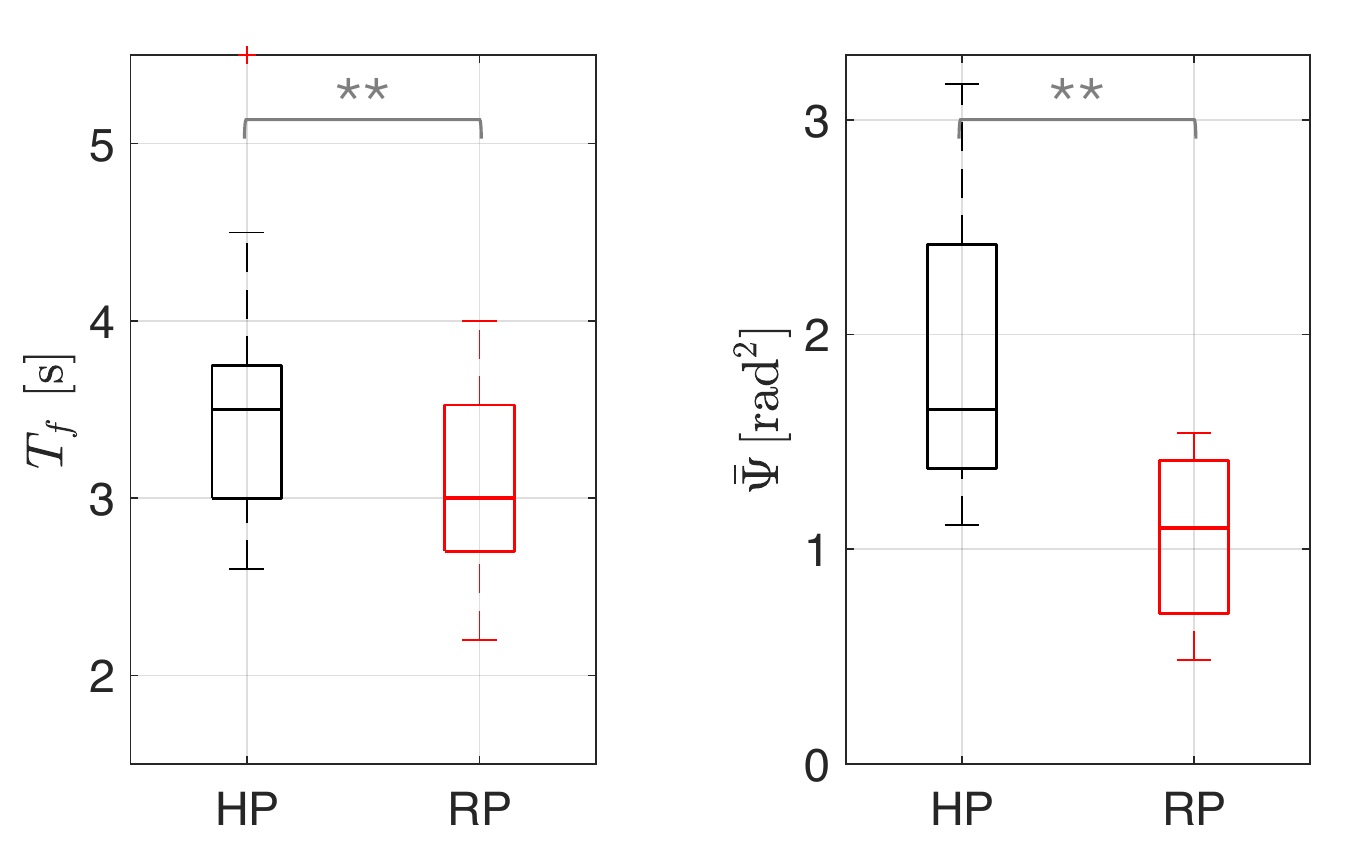}
    \vspace{-0.4cm}
    \caption{Statistical comparison between human passing (\textit{HP}) and robotic passing (\textit{RP}) conditions: (left) duration of the approach phase, indicating improved efficiency in robotic handover; (right) mean cost of compensatory movements $\bar{\Psi}$ over the approach phase, highlighting reductions in functioning joints usage. Significance levels are indicated at **$p<0.01$.} 
    \vspace{-0.3cm}
    \label{fig:metrics}
\end{figure}

The statistical analysis comparing \textit{HP} and \textit{RP} conditions across all tested $y_\text{task}$ values demonstrated that the robot handover significantly improved both the efficiency and comfort of the prosthesis user's grasp. Indeed, a significant reduction in the duration of the approach phase was observed, as shown in the left box plot of Fig. \ref{fig:metrics}. Additionally, the measured cost associated with compensatory movements, $\Psi(t_\text{interaction})$, at the interaction pose, and the mean cost representing the functioning joints usage, $\bar{\Psi}$, exhibited significant decreases, as depicted in the right box plot of Fig. \ref{fig:metrics}.
The norm of joint displacements at the wrist remained consistently close to zero in both handover conditions, as expected. However, in the \textit{HP} condition, wrist displacements occasionally reached up to $6^\circ$, indicating instances where the prosthesis user was forced to violate the wrist constraint to enable the grasp. Regarding joint motion smoothness, the integral of the squared joint jerk during the approach phase ($J$) was not presented as a box plot due to value variations introduced by the monitoring system and numerical derivation, but comparisons between the two conditions (\textit{HP} and \textit{RP}) provided consistent and meaningful insights. A significant $52.7\%$ reduction in $J$ ($p < 0.01$) in the robotic handover condition revealed smoother and more seamless grasping approaches.

\section{Discussion and Conclusions}
\label{sec:discussions_conclusions}

In this work, we presented a novel online method for mitigating compensatory movements in prosthesis users with restricted arm mobility, an often overlooked issue that can lead to long-term harm. Our approach involves creating a personalised mobility model for the prosthesis user and integrating it into a constrained optimisation framework. This framework computes the optimal user posture for task performance in a functional and comfortable manner, considering task requirements, and informs the robot, which reconfigures the task accordingly to promote the adoption of such a posture. Initial results with healthy subjects using a proposed body-powered finger prosthesis as a supernumerary finger demonstrated that a robotic assistant embedding the user-specific mobility model outperformed human partners. Improvements were observed in both the reduction of compensatory movements in functioning joints and the efficiency of the prosthesis user's grasp during handover tasks. These findings are promising for introducing collaborative robots as workplace and caregiving assistants, promoting inclusion, and facilitating the seamless integration of prosthetic devices into users' daily lives. Despite these encouraging results, we observed some deviations in how users resolved redundancy in joint usage, leading to slight differences in performance. Future works will focus on providing guidance to users on configuring their functioning joints, potentially through visual feedback interfaces developed in our simulations or vibrotactile feedback systems \cite{lorenzini2022performance}. Additionally, we plan to test the framework with prosthetic devices handling more complex amputations, such as those involving the elbow, and consider partial impairments. Finally, we aim to extend our framework to more complex interactive tasks, further evaluating its potential for enhancing prosthetic device usability in diverse real-world scenarios.


\addtolength{\textheight}{-10cm}   






\bibliographystyle{IEEEtran}

\bibliography{bibliography}

\end{document}